%% file: main.tex
\documentclass[10pt,twocolumn,letterpaper]{article}
\usepackage[pagenumbers]{cvpr} 

\input{preamble}

\usepackage{makecell}

\definecolor{cvprblue}{rgb}{0.21,0.49,0.74}
\usepackage[pagebackref,breaklinks,colorlinks,citecolor=cvprblue]{hyperref}
\usepackage{algorithmic}
\usepackage[linesnumbered,ruled,vlined]{algorithm2e}
\usepackage[toc,page]{appendix}
\usepackage{multirow}
\usepackage{pifont}
\newcommand{\cmark}{\ding{51}}
\newcommand{\xmark}{\ding{55}}
\newcommand{\tabincell}[2]{\begin{tabular}{@{}#1@{}}#2\end{tabular}}
\newcommand*{\affaddr}[1]{#1} 

\newcommand*{\email}[1]{\texttt{#1}}

\title{DREAM: Document Reconstruction via End-to-end Autoregressive Model}

\author{
	Xin Li\textsuperscript{\dag\thanks{Equal contribution. \textsuperscript{\dag}Contact person.}} \quad Mingming Gong\textsuperscript{*} \quad Yunfei Wu\textsuperscript{*} \quad Jianxin Dai \quad Antai Guo \quad Xinghua Jiang \quad \\ Haoyu Cao \quad
	Yinsong Liu \quad Deqiang Jiang \quad Xing Sun \\		
	\affaddr{Tencent YouTu Lab} \quad\\
	\email{\small \{fujikoli, riemanngong, marcowu, willsdai, ankerguo, clarkjiang, }\\
	\email{\small rechycao, jasonysliu, dqiangjiang, winfredsun\}@tencent.com}
}

\begin{document}
\maketitle
\input{0_abstract}    
\input{1_introduction}
\input{2_related_work}

\input{3_task_definition}

\input{4_methodology}

\input{5_experiments}
\input{6_conclusion}
\input{X_suppl}
{
    \small
    \bibliographystyle{ieeenat_fullname}
    \bibliography{main}
}

\end{document}

%% file: preamble.tex
\usepackage[dvipsnames]{xcolor}


%% file: 0_abstract.tex
\begin{abstract}
Document reconstruction constitutes a significant facet of document analysis and recognition, a field that has been progressively accruing interest within the scholarly community.
A multitude of these researchers employ an array of document understanding models to generate predictions on distinct subtasks, subsequently integrating their results into a holistic document reconstruction format via heuristic principles. 
Nevertheless, these multi-stage methodologies are hindered by the phenomenon of error propagation, resulting in suboptimal performance.
Furthermore, contemporary studies utilize generative models to extract the logical sequence of plain text, tables and mathematical expressions in an end-to-end process.
However, this approach is deficient in preserving the information related to element layouts, which are vital for document reconstruction.
To surmount these aforementioned limitations, we in this paper present an innovative autoregressive model specifically designed for document reconstruction, referred to as Document Reconstruction via End-to-end Autoregressive Model~(DREAM).
DREAM transmutes the text image into a sequence of document reconstruction in a comprehensive, end-to-end process, encapsulating a broader spectrum of document element information.
In addition, we establish a standardized definition of the document reconstruction task, and introduce a novel Document Similarity Metric~(DSM) and DocRec1K dataset for assessing the performance of the task.
Empirical results substantiate that our methodology attains unparalleled performance in the realm of document reconstruction.
Furthermore, the results on a variety of subtasks, encompassing document layout analysis, text recognition, table structure recognition, formula recognition and reading order detection, indicate that our model is competitive and compatible with various tasks.
\end{abstract}

%% file: 1_introduction.tex
\section{Introduction}
\begin{figure}[htb]
	\centering
	\includegraphics[width=1\linewidth, height=8cm]{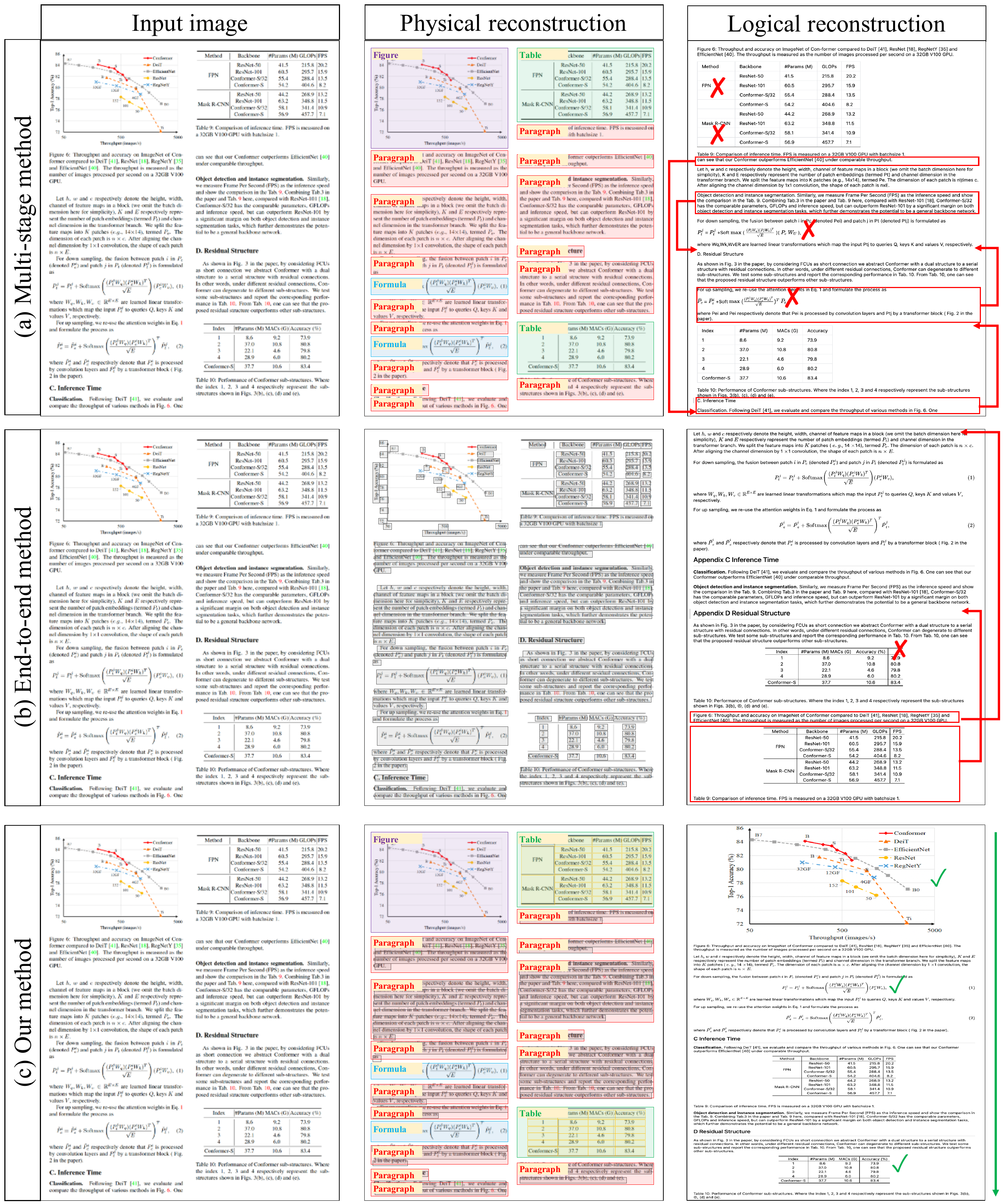}
	\caption{{Illustration of motivation of the proposed DREAM. ~(a) The multi-stage based approach, utilizes a variety of document understanding models to predict distinct document elements, subsequently amalgamating them into a comprehensive document reconstruction format via heuristic rules. However, this method is constrained by the propagation of performance errors. ~(b) The end-to-end based approach, which employs generative models to acquire the sequence of plain text, tables and mathematical expressions, yet falls short in preserving the physical information pertaining to layouts and performs not satisfactorily enough on the logical reconstruction result. ~(c) Our proposed DREAM, which exhibits the capability to restore an extensive range of document contents in a more holistic manner, encompasses physical and logical reconstruction information.}} 
	\label{fig:motivation}
\end{figure}
Document reconstruction is an integral research of document understanding, which harbors significant potential to provide richer structured information for Large Language Models~(LLMs) and enhances their capacity to interpret image data.
This endeavor seeks to scrutinize disparate elements~(\textit{e.g.}, paragraphs, tables, formulas and figures) within document images, and subsequently restore them to their original structures in both \textit{physical} and \textit{logical} formats.
More concretely, \textit{physical reconstruction}~\cite{zhong2019publaynet, pfitzmann2022doclaynet, li2020docbank, cheng2023m6doc} pertains to the visual representation of the document, which differentiates regions characterized by distinct types of elements, inclusive of the categories and bounding boxes of various elements.
Conversely, \textit{logical reconstruction}~\cite{liu2020abcnet, lyu2018mask, huang2022swintextspotter, jaume2019funsd, huang2019icdar2019, park2019cord, gobel2013icdar, chi2019complicated, zhong2020image, li2020tablebank, liu2021show, nassar2022tableformer, liu2022neural, huang2023improving, mouchere2014icfhr, deng2017image, yuan2022syntax, wang2021layoutreader, blecher2023nougat, lv2023kosmos} discerns the semantic structures of documents based on their inherent meaning and assigns them to various semantic information, consisting of the transcription of plain text, the markup language of table and the symbols of formula.
This complex task encompasses a multitude of subtasks, including but not limited to, document layout analysis~\cite{zhong2019publaynet, pfitzmann2022doclaynet, li2020docbank, cheng2023m6doc}, text recognition~\cite{liu2020abcnet, lyu2018mask, huang2022swintextspotter, jaume2019funsd, huang2019icdar2019, park2019cord}, table structure recognition~\cite{gobel2013icdar, chi2019complicated, zhong2020image, li2020tablebank, liu2021show, nassar2022tableformer, liu2022neural, huang2023improving}, formula recognition~\cite{mouchere2014icfhr, deng2017image, yuan2022syntax} and reading order detection~\cite{wang2021layoutreader}.
To date, numerous algorithms have demonstrated remarkable advancements by leveraging deep learning techniques within the community.
Nonetheless, the task remains a formidable challenge due to the extensive diversity in document layouts and presentation styles.

One category of prevalent document reconstruction methodologies~\cite{zhong2019publaynet, pfitzmann2022doclaynet, li2020docbank, cheng2023m6doc, liu2020abcnet, lyu2018mask, huang2022swintextspotter, jaume2019funsd, huang2019icdar2019, park2019cord, gobel2013icdar, chi2019complicated, zhong2020image, li2020tablebank, liu2021show, nassar2022tableformer, liu2022neural, huang2023improving, mouchere2014icfhr, deng2017image, yuan2022syntax, wang2021layoutreader, tang2022few, liu2023spts, tang2022optimal, zhao2024multi, zhao2025tabpedia, feng2023unidoc, tang2022youcan} involves the prediction of multiple subtasks, which are subsequently concatenated to form a pipeline, as shown in Fig.~\ref{fig:motivation}(a).
These approaches are proficient in one or several domain-specific subtasks, such as document layout analysis, text recognition and table structure recognition.
To achieve a more holistic structure, the individual outcomes of various models, each addressing different subtasks, are sequentially integrated to recover the complete structure in the form of heuristic rules.
Despite the ability of these multi-stage methodologies to restore the comprehensive structure, their performance is often hindered by the multi-stage transfer loss.

To circumvent this constraint, an alternative approach, known as the end-to-end method~\cite{blecher2023nougat,lv2023kosmos}, has garnered increasing scholarly attention within the field of document reconstruction, as depicted in Fig.~\ref{fig:motivation}(b).
For instance, Nougat~\cite{blecher2023nougat} employs Donut~\cite{kim2022ocr} to transform scientific documents into a markup language, thereby providing a promising solution to enhance the accessibility of scientific knowledge in the digital era.
KOSMOS-2.5~\cite{lv2023kosmos} introduces a multimodal large language model to transmute images into spatially-aware texts and markdown-formatted texts.
These recent advancements offer a promising end-to-end solution to enhance the capabilities of document reconstruction.
However, they often fall short in preserving the physical information pertaining to layouts and perform not satisfactorily enough on the logical reconstruction result, which is integral to restoring the complete structure.

In this work, we introduce an innovative end-to-end document reconstruction methodology to overcome the above limitations, designated as Document Reconstruction via End-to-end Autoregressive Model~(DREAM), as illustrated in Fig.~\ref{fig:motivation}(c).
Concretely, we define the output form of document reconstruction to standardize the definition of the task.
In this way, the model transforms an image into a sequence of document reconstruction in an end-to-end manner, which effectively addresses the issue of accuracy loss engendered by the multi-stage methods utilized in preceding methodologies.
In contrast to prior strategies, our method is capable of restoring a wide array of document contents in a more comprehensive manner  incorporating both logical and physical information.
Furthermore, we also introduce a new Document Similarity Metric~(DSM) and DocRec1K dataset to assess the performance of the task.
In summary, our primary contributions are in the three folds:
\begin{itemize}
	\item We standardize the definition of document reconstruction task incorporating both physical and logical information, and introduce a new  DSM metric and DocRec1K dataset for evaluating its performance. To the best of our knowledge, this is the first research to investigate the definition and evaluation of a more comprehensive document reconstruction structure.
	\item We propose a novel document reconstruction model, termed DREAM, which transforms the text image into a sequence of document reconstruction in an end-to-end autoregressive manner, encompassing a more comprehensive array of document element information.
	\item Experimental results substantiate that our methodology achieves state-of-the-art performance on the document reconstruction task. Moreover, the results on various subtasks across different benchmarks also indicate that our model is comparable and compatible with diverse tasks.
\end{itemize}

%% file: 2_related_work.tex
\section{Related Work}
\begin{figure*}[htp]
	\centering
	\includegraphics[width=1\linewidth]{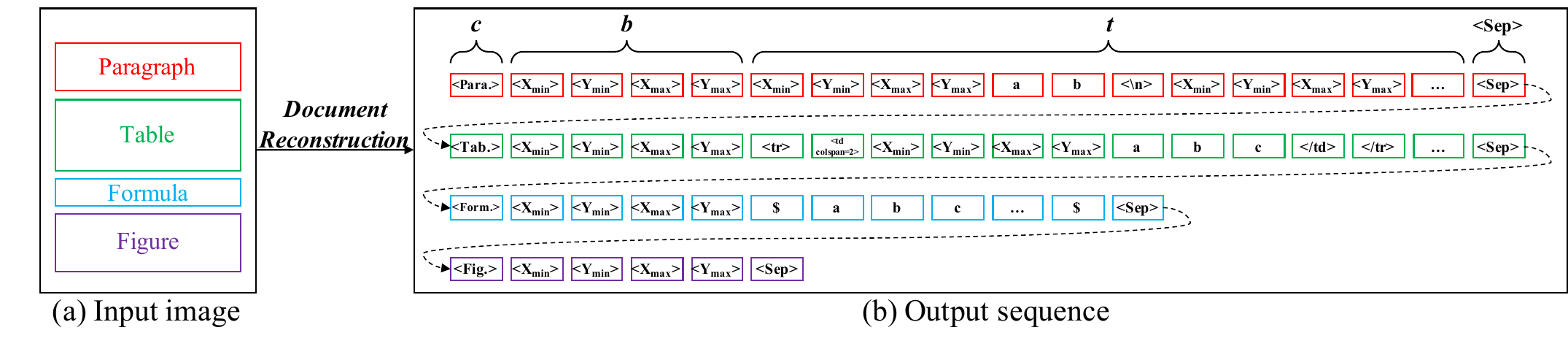}
	\caption{The task definition of document reconstruction.}
	\label{fig:defination}
\end{figure*}
\subsection{Multi-stage based Document Reconstruction}
The process of multi-stage document reconstruction~\cite{zhong2019publaynet, pfitzmann2022doclaynet, li2020docbank, cheng2023m6doc, liu2020abcnet, lyu2018mask, huang2022swintextspotter, jaume2019funsd, huang2019icdar2019, park2019cord, gobel2013icdar, chi2019complicated, zhong2020image, li2020tablebank, liu2021show, nassar2022tableformer, liu2022neural, huang2023improving, mouchere2014icfhr, deng2017image, yuan2022syntax, wang2021layoutreader} is traditionally divided into a pipeline of distinct stages.
These include: 1) analyzing the layout elements of the document, which determine the position and category of texts, images and other elements within a document image, 2) arranging these elements in a sequential order, enabling the presentation of a complete document content in a logical and seamless reading sequence, 3) employing specific methods to extract the textual content from the respective layout elements, such as text recognition, table structure recognition and formula recognition.
A more detailed description is given in supplementary material.

However, multi-stage based document reconstruction optimizes each stage independently, potentially resulting in suboptimal outcomes. 
Errors from one stage will propagate and amplify in subsequent stages.
Conversely, end-to-end systems are less susceptible to such error propagation as they learn to optimize for the final output directly.

\subsection{End-to-end based Document Reconstruction}
The advent of generative models has opened up new possibilities for end-to-end document reconstruction~\cite{blecher2023nougat,lv2023kosmos}.
Current prevalent strategies, such as those employed by Nougat~\cite{blecher2023nougat} and KOSMOS-2.5~\cite{lv2023kosmos}, utilize generative transformers to sequentially generate the markdown information of different layout elements following the reading order.
During the training process, Nougat~\cite{blecher2023nougat} exclusively used PDF data, while KOSMOS-2.5~\cite{lv2023kosmos} broadened the scope of training data to include README, DOCX, and HTML documents. 
Furthermore, KOSMOS-2.5~\cite{lv2023kosmos} introduced a cooperative transcription task that generates spatially aware text blocks, in addition to generating markdown format texts, which operates under a shared vision encoder and task-specific text decoder to enhance the document reconstruction presented by markdown text and facilitate a unified multimodal literate capability.

The end-to-end approach is advantageous as it enables the model to comprehend the context and structure of the layout, which is anticipated to yield more accurate reconstruction results.
Unfortunately, the above end-to-end technologies lead to incomplete content reconstruction, as they overlook the physical information and show unsatisfactory performance on the logical sequence of transcription.
The proposed DREAM effectively addresses the limitations of existing technical solutions and achieves complete document reconstruction in an end-to-end manner.

%% file: 3_task_definition.tex
\section{Task, Metric and Dataset}
\subsection{Task Definition} \label{sec:definition}
As depicted in Fig.~\ref{fig:defination}, the task of document reconstruction aims to transform the document image into a sequence of physical and logical information, which concurrently localizes a variety of element types and identifies their contents.
Each layout element can be defined by three distinct components: 1) the category $c$, which is allocated a unique token for each type of element, such as $\textless {\rm Paragraph} \textgreater$, $\textless {\rm Table} \textgreater$, $\textless {\rm Formula} \textgreater$, $\textless {\rm Figure} \textgreater$, 2) the bounding box $b$, represented by a quartet of tokens $\left\lbrace \textless {\rm X_{min}} \textgreater, \textless {\rm Y_{min}} \textgreater, \textless {\rm X_{max}} \textgreater, \textless {\rm Y_{max}} \textgreater \right\rbrace$ to denote the top-left and bottom-right coordinates of the element within the image, and 3) the transcription content $t$, which adheres to a specific text format of variable length.
Thus, the sequence of an element can be expressed as $\mathbf{y} = \left\lbrace c, b, t, \textless  {\rm Sep} \textgreater \right\rbrace$, where the $\textless {\rm Sep}\textgreater$ token is appended to the end of the element to distinguish different elements within the reconstructed text.
As a result, the comprehensive sequence of the document reconstruction outcome, which contains multiple elements in strict accordance with the layout reading order, is presented as $\mathbf{Y} = {\left\lbrace \mathbf{y}^{k} \right\rbrace}_{k=1}^{K}$, where $K$ signifies the number of elements.

Given the distinct semantic structures inherent to diverse categories of document elements, we strategically leverage a series of unique tokens to encapsulate the transcription content $t$. 
Specifically, the content within the element $\textless {\rm Paragraph} \textgreater$ may encompass multiple lines of unformatted text, which are considered sub-elements.
Rather than employing the unformatted text to symbolize the text lines within a paragraph, we utilize the content coordinates and characters to represent the sub-element line, subsequently incorporating the token $\textless {\rm \backslash n} \textgreater$ to demarcate various lines from each other, as depicted by the red boxes in Fig.~\ref{fig:defination}(b).
Furthermore, we engage HTML representations to denote the transcription outcomes of the $\textless {\rm Table} \textgreater$, which includes tokens such as $\textless {\rm tr} \textgreater$, $\textless {\rm /tr} \textgreater$, $\textless {\rm td} \textgreater$, $\textless {\rm /td} \textgreater$, in conjunction with the attributes ``rowspan'' and ``colspan'', as illustrated by the green boxes in Fig.~\ref{fig:defination}(b).
In addition, the coordinates $b$ are integrated into the HTML codes to represent the bounding box of each cell within a table similar to the line processing for paragraphs.
To articulate the category of $\textless {\rm Formula} \textgreater$, we adopt the LaTeX language, represented as the blue information in Fig.~\ref{fig:defination}(b), to annotate the contents and furnish a plethora of mathematical symbols and commands that cater to the requirements of intricate formulas.
Notably, the transcription of $\textless {\rm Figure} \textgreater$ is simply denoted as empty, as indicated by the purple contents in Fig.~\ref{fig:defination}(b).

\subsection{Evaluation Metric}  \label{sec:metric}
In recent literature pertaining to document analysis and recognition, three primary categories of metrics are predominantly employed to assess the efficacy of document reconstruction: mean Average Precision (mAP), Edit Distance (ED), and subtask-specific instruments.
Specifically, mAP is utilized as a metric to gauge the accuracy of element categories and locations within the context of layout analysis tasks.
This metric emphasizes the physical information inherent in document reconstruction, yet it does not adequately represent the specific texts and the layout ordering of document objects.
An alternative set of evaluation protocols, encompassing Normalized Edit Distance (NED) and Normalized Tree Edit Distance (NTED), has been recently adopted in the study~\cite{lv2023kosmos}.
These metrics evaluate the quality of image-to-markdown generation, however, they do not effectively measure the layout information of various document elements.
Furthermore, a range of methods introduces diverse specific metrics for different subtasks of document reconstruction, such as the F1-score of words for text recognition, TEDS~\cite{zhong2020image} for table structure recognition, and so forth.
These metrics concentrate on a particular aspect of document understanding, rendering them inadequate for evaluating the overall effectiveness of document reconstruction.
In conclusion, despite the specific focus of previous metrics, it remains a formidable challenge to holistically assess the performance of the document reconstruction task.

To more effectively assess the comprehensive performance of document reconstruction, we present an innovative Document Similarity Metric~(DSM) for this task, which simultaneously evaluates the physical and logical information of document elements.
Drawing inspiration from the concept of edit distance, we incorporate the location cost and the transcription cost into the distance computation between two elements, as opposed to solely considering the isolated string distance cost.
Let $K$ represents the actual number of elements within a document image, and $\widetilde{K}$ denotes the predicted number of elements.
Assuming the category and coordinates of the $i$-th ground truth element are $c_{i}$ and $b_{i}$, those of the $j$-th predicted element are $\widetilde{c}_{j}$ and $\widetilde{b}_{j}$, the location cost between the two elements is defined as:
\begin{equation}
	\small
	\begin{aligned}
	Cost_{loc.}(i, j) = [\underbrace{1_{\left\lbrace c_{i} \neq \widetilde{c}_{j} \right\rbrace}}_{category} + \underbrace{1 - { IoU}(b_{i}, \widetilde{b}_{j})}_{coordinates}] / 2.
	\end{aligned}
\end{equation}
In this equation, the left part signifies that the cost is 0 when the categories are identical while 1 when they differ, and the right part represents the non-overlapping degree between two bounding boxes utilizing the function of ${IoU}$.
As for the transcription cost, it indicates a string-based comparison between the ground truth ${t_{i}}$ and predicted $\widetilde{t}_{j}$, akin to the edit distance function, which is formulated as:
\begin{equation}
	\small
	\begin{aligned}
		Cost_{tran.}(i, j) = \underbrace{Dist\left(t_i, \widetilde{t}_j\right) / { Maxlen} \left(t_i, \widetilde{t}_j\right)}_{transcription}.
	\end{aligned}
\end{equation}
Here, $Dist$ and ${ Maxlen}$ symbolize the edit distance and the maximum length between two strings, respectively.
Subsequently, the distance between the prediction element and the ground truth element is computed as:
\begin{equation}
	\small
	\begin{aligned}
		Cost(i, j) = [Cost_{loc.}(i, j) + Cost_{tran.}(i, j)] / 2,
	\end{aligned}
\end{equation}
which averages the outcomes of $Cost_{loc.}$ and $Cost_{tran.}$.
To further examine the performance of the entire document reconstruction, consisting of the ground truth sequence from $\mathbf{y}^{1}$ to $\mathbf{y}^{i}$ and the predicted sequence from $\mathbf{\widetilde{y}}^{1}$ to $\mathbf{\widetilde{y}}^{j}$, we accumulate the costs of elements using dynamic programming:
\begin{equation}
	\small
	\begin{aligned}
		D(i, j) &= {Min}(D(i-1, j), D(i, j-1), D(i-1, j-1)) \\
		&+ Cost(i, j).
	\end{aligned}
\end{equation}
Consequently, the DSM is formulated as:
\begin{equation}
	\small
	\begin{aligned}
		\textit{DSM} = 1-\frac{1}{B} \sum_{b=1}^B D(K^{b}, \widetilde{K}^{b}) / { Maxlen}(K^{b}, \widetilde{K}^{b}).
	\end{aligned}
\end{equation}
In this formula, $B$ denotes the number of document images, $D(K^{b}, \widetilde{K}^{b})$ indicates the distance of the $b$-th reconstruction, and is normalized by the maximum sequence length. The value of \textit{DSM} ranges from 0 to 1, with a higher value signifying that the prediction is closer to the ground truth.

\begin{figure*}[htb]
	\centering
	\includegraphics[width=1\linewidth]{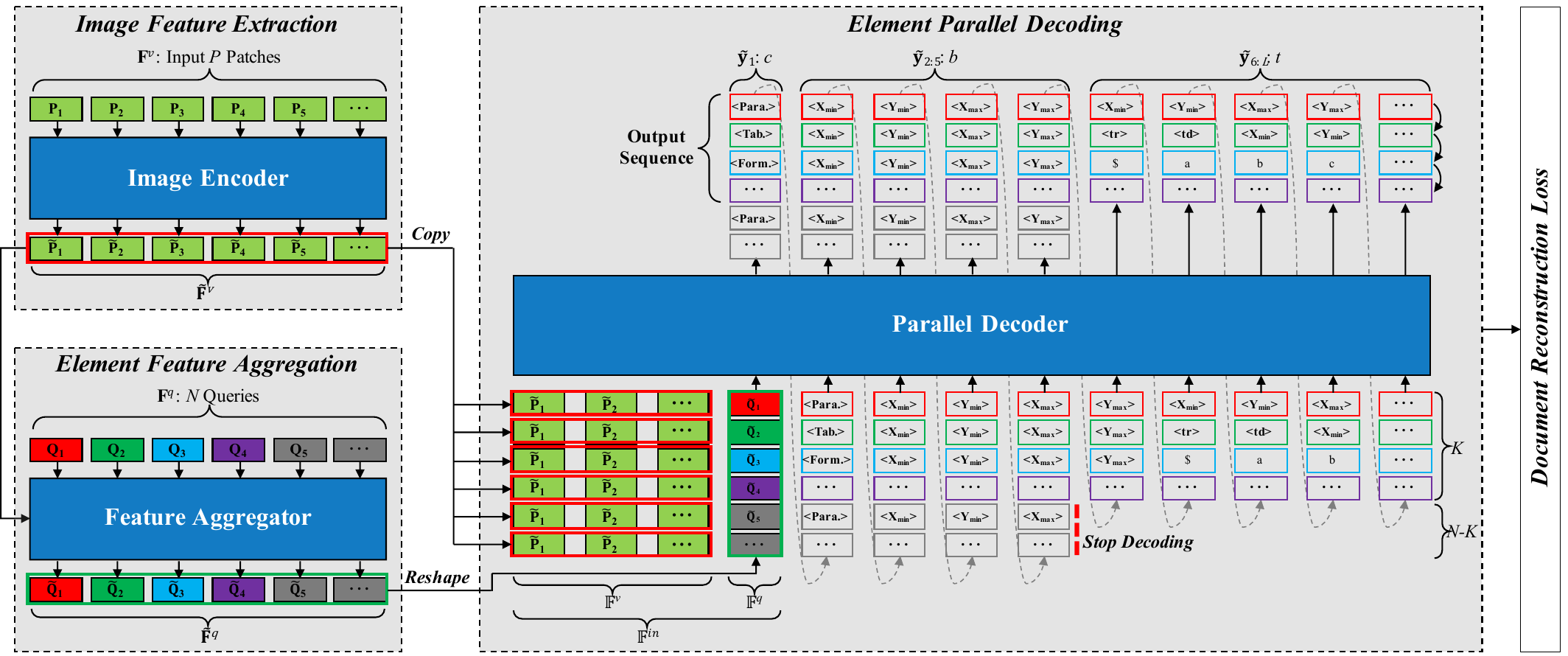}
	\caption{The architecture of our proposed DREAM, in which the gray color of boxes denotes the negative queries to be filtered while the other colors represent the actual elements retained. The model is optimized in an end-to-end parallel scheme.}
	\label{fig:architecture}
\end{figure*}

\subsection{Evaluation Dataset}  \label{sec:dataset}
The extant public repositories of data pertaining to document reconstruction fail to sufficiently address the particular task delineated herein.
As a result, we construct a unique dataset, referred to as \textbf{DocRec1K}, to assess the efficacy of the document reconstruction task.
Acknowledging the complexities inherent in replicating layout data annotation, we elect to construct this dataset on the basis of DocLayNet~\cite{pfitzmann2022doclaynet}, eschewing the need to initiate the process from the ground up. 
DocLayNet is distinguished by its superior layout element annotation, a product of meticulously devised annotation rules and a scientifically rigorous annotation procedure, which significantly reduces the cost associated with the creation of comprehensive document layout reconstruction datasets.
Consequently, the construction process of DocRec1K is bifurcated into three distinct stages: 1) the sequencing of layout elements, 2) the association of these elements with their corresponding textual content, and 3) the conversion of element contents into the specific formats previously delineated.
A more comprehensive elucidation is provided in the supplementary material.

%% file: 4_methodology.tex
\section{Methodology}

\subsection{Overall Architecture}
The overview of the proposed DREAM is shown in Fig.~\ref{fig:architecture}, which utilizes an end-to-end architecture to reconstruct the document structure, eliminating the necessity for any intermediary pipelines.
The model is primarily composed of three components: Image Encoder~(IE), Feature Aggregator~(FA) and Parallel Decoder~(PD).
Initially, the input image undergoes a transformation into a patch embedding, followed by feature extraction executed by IE, resulting in a sequence of image tokens.
Subsequently, the element queries, in conjunction with the image features, are introduced into FA.
This aggregator employs cross-attention mechanism to stimulate the element queries to aggregate information from the image.
In the final stage, the element queries along with image tokens are directed to PD. 
This component autoregressively generates comprehensive information for each element in a parallel fashion, including the category, coordinates and texts of each element. 
The optimization of DREAM is achieved through the minimization of the \textit{Document Reconstruction Loss}.

\subsection{Image Feature Extraction}
In this stage, the input image, denoted as $\mathbf{x}$, is transformed to a sequence of patch embeddings, represented as $\mathbf{F}^{v} =  {\left\lbrace {\mathbf{{P}}_{i}} \right\rbrace}_{i=1}^{P} \in \mathbb{R}^{P \times d}$.
Here, $P$ signifies the patch size of the image and $d$ is the dimension of the patch.
Subsequently, the image encoder Swin~\cite{liu2021swin} is utilized to process these patches into a collection of aggregated image embeddings, symbolized as $\mathbf{\widetilde{F}}^{v} =  {\left\lbrace {\mathbf{\widetilde{P}}_{i}} \right\rbrace}_{i=1}^{P} \in \mathbb{R}^{P \times d}$.

\subsection{Element Feature Aggregation}
Previous document reconstruction methodologies have attempted to directly forecast the comprehensive markup text from the document image.
Unfortunately, the sequence produced by such an approach tends to be excessively lengthy and susceptible to recurrent generation, not to mention the considerable time consumption.
To address these issues, we introduce a novel parallel decoding paradigm that simultaneously generates sequences of disparate elements, as opposed to predicting in a purely autoregressive fashion.
More specifically, we incorporate the element query mechanism to perform feature aggregation of document elements, subsequently generating the aggregated element representations in parallel.
Motivated by DETR~\cite{carion2020end}, we design a feature aggregator that amalgamates the representations of document elements.
This aggregator functions by generating context features for each element, utilizing a cross-attention mechanism.
During the phase of element feature aggregation, $N$ element queries $\mathbf{F}^{q} = {\left\lbrace {\mathbf{{Q}}_{i}} \right\rbrace}_{i=1}^{N} \in \mathbb{R}^{N \times d}$ act as the query, while the image features $\mathbf{\widetilde{F}}^{v}$ are employed as the key and value.
Subsequent to this procedure, the contextual image information is embedded into the element queries as $\mathbf{\widetilde{F}}^{q} = {\left\lbrace {\mathbf{\widetilde{Q}}_{i}} \right\rbrace}_{i=1}^{N} \in \mathbb{R}^{N \times d}$.

\subsection{Element Parallel Decoding}
Upon the aggregation of elemental features, our goal is to parallelly generate the information pertaining to each element, which encompasses type, coordinates and texts as delineated in Sec.~\ref{sec:definition}.
Specifically, the visual tokens $\mathbf{\widetilde{F}}^{v}$ are initially duplicated into $N$ copies, denoted as $\mathbb{F}^{v} = {\left\lbrace \mathbf{\widetilde{F}}^v_{i} \right\rbrace}_{i = 1}^{N} \in \mathbb{R}^{N \times P \times d}$, and $\mathbf{\widetilde{F}}^{q}$ is reshaped from the shape of $N \times d$ to $N \times 1 \times d$, resulting in $\mathbb{F}^{q} \in \mathbb{R}^{N \times 1 \times d}$. 
Subsequently, the $N$ element features $\mathbb{F}^{q}$ are appended to the visual features $\mathbb{F}^{v}$, serving as the decoding input $\mathbb{F}^{in} \in \mathbb{R}^{N \times (P+1) \times d}$ for the decoder.
During the decoding stage, the parallel decoder first generates five tokens of the sequence autoregressively, representing the category and discretized coordinates of a single element.
It is important to note that these $N$ elements comprise positive and negative cases, necessitating the filtration of negative examples to procure valid elements and circumvent the time-intensive generation of all elements.
Thanks to the mechanism of bipartite matching~\cite{carion2020end}, our end-to-end method can achieve significant speed by filtering out elements with the confidence below 80\%, as opposed to the method~\cite{blecher2023nougat,lv2023kosmos} that greedily employs a large number of redundant elements to transcribe their details.
Given the actual number of elements of interest in a document image is $K$, 
the corresponding decoded tokens are $\mathbf{\widetilde{Y}}_{1}^{5} = {\left\lbrace \mathbf{\widetilde{y}}_{1:5}^{k} \right\rbrace}^{K}_{k = 1} \in \mathbb{R}^{K \times 5}$, where $\mathbf{\widetilde{y}}_{1:5}^{k}$ signifies the tokens of class $c_{k}$ along with coordinates $b_{k}$ for the $k$-th element, and $\mathbf{\widetilde{Y}}_{1}^{5}$ represents the predicted physical information of the $K$ elements.
Subsequently, these retained elements continue autoregressively decoding to predict the text tokens $\mathbf{\widetilde{Y}}^{L}_{6} =  {\left\lbrace\mathbf{\widetilde{y}}_{6:L}^{k}\right\rbrace}_{k=1}^{K}  \in \mathbb{R}^{K \times (L - 5)}$ within the element until the end of the sequence token occurs, where $\mathbf{\widetilde{y}}_{6:L}^{k}$ denotes the logical transcription texts of the $k$-th element, and $L$ is the maximum of the predicted sequences.
Furthermore, the complete sequences of the valid elements can be obtained, which concatenate $\mathbf{\widetilde{Y}}_{1}^{5}$ and $\mathbf{\widetilde{Y}}_{6}^{L}$ to $\mathbf{\widetilde{Y}}^{L}_{1} \in \mathbb{R}^{K \times L}$.
Ultimately, these parallel generated element results are amalgamated with the padding tokens removed to procure the final sequence of document reconstruction.

\subsection{Document Reconstruction Loss}
In an endeavor to optimize DREAM in an end-to-end manner, we propose an innovative \textit{Document Reconstruction Loss}, which is composed of three distinct components: element discrimination loss, element transcription loss and sequence reconstruction loss.

\noindent\textbf{Element discrimination loss:}
To extract the $K$ elements from the larger predefined $N$ predictions, we adopt the methodology from previous work~\cite{carion2020end}, utilizing the Hungarian algorithm for the element discrimination loss $\mathcal{L}_{ed}$ to establish a match between the predicted and ground truth elements.
Furthermore, the cross-entropy loss is employed to optimize the preceding tokens.
We designate the ground truth tokens of $K$ elements are $\mathbf{{Y}}_{1}^{L} =  {\left\lbrace\mathbf{{y}}_{1:L}^{k}\right\rbrace}_{k=1}^{K}  \in \mathbb{R}^{K \times L}$, wherein the tokens of each element $\mathbf{{y}}_{1:L}^{k}$ are padding to a fixed length $L$ and masked by a matrix $M_{1}^{L}$. 
Given that the ground truth class and coordinates of the $n$-th element are $c^{n} = \mathbf y^{n}_{1}$ and $b^{n} = \mathbf y^{n}_{2:5}$ respectively, we define the probability of class $c^{n}$ as $\widetilde{P}_{\widetilde{\sigma}(n)}$ and the predicted coordinates as $\widetilde{b}^{\widetilde {\sigma}(n)}$, where $\widetilde{\sigma}$ is the optimal assignment computed by the bipartite matching between the predictions and the ground truth set of objects.
In this way, the loss $\mathcal{L}_{ed}$ is defined as:
\begin{equation}
	\small
	\begin{aligned}
		\mathcal{L}_{ed} = &\sum_{n=1}^N (- log  \widetilde{P}_{\widetilde{\sigma}(n)}(c^{n}) + {IoU}(\widetilde{b}^{\widetilde {\sigma}(n)}, b^{n})) \\
		&- \sum_{k=1}^{K} \sum_{t=1}^{5} log P(\mathbf{\widetilde{y}}_{t}^{k}|\left\lbrace\mathbb{F}^{in}, \mathbf y_{1:(t-1)}^{k}\right\rbrace)  \cdot M_{1}^{5},
	\end{aligned}
\end{equation}
where $\left\lbrace \mathbb{F}^{in}, \mathbf y_{1:(t-1)}^{k} \right\rbrace$ and $\mathbf{\widetilde{y}}_{t}^{k}$ denote the input and output of the $k$-th element for the $t$-th decoding step respectively.

\noindent\textbf{Element transcription loss:} This loss is utilized to optimize the transcription texts of the retained $K$ elements with the cross-entropy loss, which is defined as:
\begin{equation}
	\small
	\begin{aligned}		
		\mathcal{L}_{et} = - \sum_{k=1}^{K} \sum_{t=6}^{L}logP(\mathbf{\widetilde{y}}_{t}^{k}|\left\lbrace\mathbb{F}^{in}, \mathbf y_{1:(t-1)}^{k}\right\rbrace) \cdot M_{6}^{L}
	\end{aligned}
\end{equation}

	\noindent\textbf{Sequence reconstruction loss:} To ensure the reading order of elements within the document, the sequence reconstruction loss is applied to DREAM, which employs the cosine similarity loss to measure the discrepancy between the entire sequence of input and output:
\begin{equation}
	\small
	\begin{aligned}
		\mathcal{L}_{sr} = 1 - \frac{{{ flat}(\mathbf{\widetilde{Y}}_{1}^{L} \cdot M_{1}^{L})} \cdot { flat}(\mathbf{Y}_{1}^{L} \cdot M_{1}^{L})}{|| {{ flat}(\mathbf{\widetilde{Y}}_{1}^{L} \cdot M_{1}^{L})} || \cdot || { flat}(\mathbf{Y}_{1}^{L} \cdot M_{1}^{L}) ||},
	\end{aligned}
\end{equation}
where ${ flat}$ signifies the process of flattening the results of multiple elements into a single dimension.

As a result, the global optimization can be defined as:
\begin{equation}
	\small
	\begin{aligned}
		\mathcal{L} = \lambda_{1} \mathcal{L}_{ed} + \lambda_{2} \mathcal{L}_{et} + \lambda_{3} \mathcal{L}_{sr}, 
	\end{aligned}
\end{equation}
where $\lambda_{1}$, $\lambda_{2}$ and $\lambda_{3}$ represent the weight parameters used to amalgamate the multi-task losses.

%% file: 5_experiments.tex
\section{Experiments}

\subsection{Training Datasets}
Our DREAM is trained on a rich array of datasets from diverse sources, which consists of 10 million image and ground truth pairs.
DREAM is trained to utilize approximately 10 million pairs of images and corresponding ground truth data.
The open-access research-sharing platform arXiv provides the primary LaTeX source codes, then are processed to the images and annotations.
This section provides a comprehensive description of the generation of the images and ground truth.

\noindent\textbf{Image:}
The collected LaTeX source codes are compiled into PDF format using a TEX compiler, and subsequently rendered into images suitable for training purposes.

\noindent\textbf{Physical structure:}
In order to obtain the physical information, which includes categories and coordinates of various elements, we employ LaTeXML and PDFMiner tools to process the compiled PDFs.
Specifically, LaTeXML is utilized to convert the PDFs into XML representations, wherein the XML trees comprise different categories of text contents.
Furthermore, PDFMiner is employed to transform the PDFs into layout analysis results, encompassing the categories and coordinates of diverse elements.
Notably, for the sub-elements such as the multiline text within the paragraph and cells within the table, PDFMiner can only handle the former.
As a result, we additionally incorporate SLANet~\cite{li2022pp} from the PaddleOCR~\cite{du2020pp} tool to extract the coordinates of cells.
Subsequently, a fuzzy string matching algorithm, in conjunction with intersection-over-union, inspired by PubLayNet~\cite{zhong2019publaynet}, is employed to match these results and obtain the physical structures.

\noindent\textbf{Logical structure:}
To ascertain the logical information of various elements, we adhere to the processing in Nougat~\cite{blecher2023nougat} for handling the LaTeX source codes.
The source files are converted into HTML format using LaTeXML and subsequently parsed into the Markdown files to derive the logical structures.
Note that the transcriptions of the table are denoted by the HTML tags rather than Markdown formats.

Upon processing both the physical and logical structures, we employ the fuzzy string matching algorithm to match the results, ultimately generating the training ground truth sequences.

\subsection{Evaluation Datasets and Protocols}
The assessment is conducted on six distinct categories of tasks pertinent to document analysis and recognition: 1) The document reconstruction task on DocRec1K, as established in Sec.~\ref{sec:dataset}, is evaluated using the DSM delineated in Sec. ~\ref{sec:metric} and the widely accepted NED, 2) The document layout analysis task on DocBank~\cite{li2020docbank}, PubLayNet~\cite{zhong2019publaynet} and DocLayNet~\cite{pfitzmann2022doclaynet}, employs mAP as a measurement tool, 3) The text recognition task on FUNSD~\cite{jaume2019funsd}, SROIE~\cite{huang2019icdar2019} and CORD~\cite{park2019cord} utilizes the F1-score for evaluation, 4) The table structure recognition task on ICDAR-2013~\cite{gobel2013icdar}, SciTSR~\cite{chi2019complicated} and TableBank~\cite{li2020tablebank} leverages F1-score/BLEU as the evaluative metrics, 5) The formula recognition task on IM2LATEX-100K~\cite{deng2017image} and CROHME-2014~\cite{mouchere2014icfhr} is assessed under the BLEU protocol, and 6) The reading order detection task on ReadingBank~\cite{wang2021layoutreader} applies the BLEU.
A more detailed description is provided in supplementary material.

\subsection{Implementation Details}
\noindent\textbf{Experimental setting:}
We employ Swin-B~\cite{liu2021swin} as our visual encoder, configured with layer numbers \{2, 2, 14, 2\} and a window size of 10. For the parallel decoder, we extract the initial four layers from BART~\cite{lewis2019bart}. The model is initialized using pre-trained weights. 
We set the input resolution at 1024$\times$1024. 
Document images are first rescaled and subsequently padded to meet the specified input size. 
The amount of queries $N$ is set to 200 in the feature aggregator module.
The maximum sequence length $L$ for the parallel decoder is established at 1536. 
For the training loss, we empirically set all weight parameters $\lambda_{1}$ = $\lambda_{2}$ = $\lambda_{3}$ = 1.
The experiments are conducted with a total batch size of 32.
We utilize the AdamW optimizer~\cite{kingma2014adam} and train the model over 10 epochs.
The learning rate is set at $3e^{-5}$, which gradually decreases to zero by the end of the training phase.

\noindent\textbf{Post-processing on the predicted results:}
To evaluate our model on various tasks, we perform lightweight post-processing on the results output by DREAM to convert them to the corresponding formats of different tasks.
A more detailed description is given in supplementary material.

\subsection{Performance}
   \begin{figure*}[t]
	\centering
	\includegraphics[width=1\linewidth]{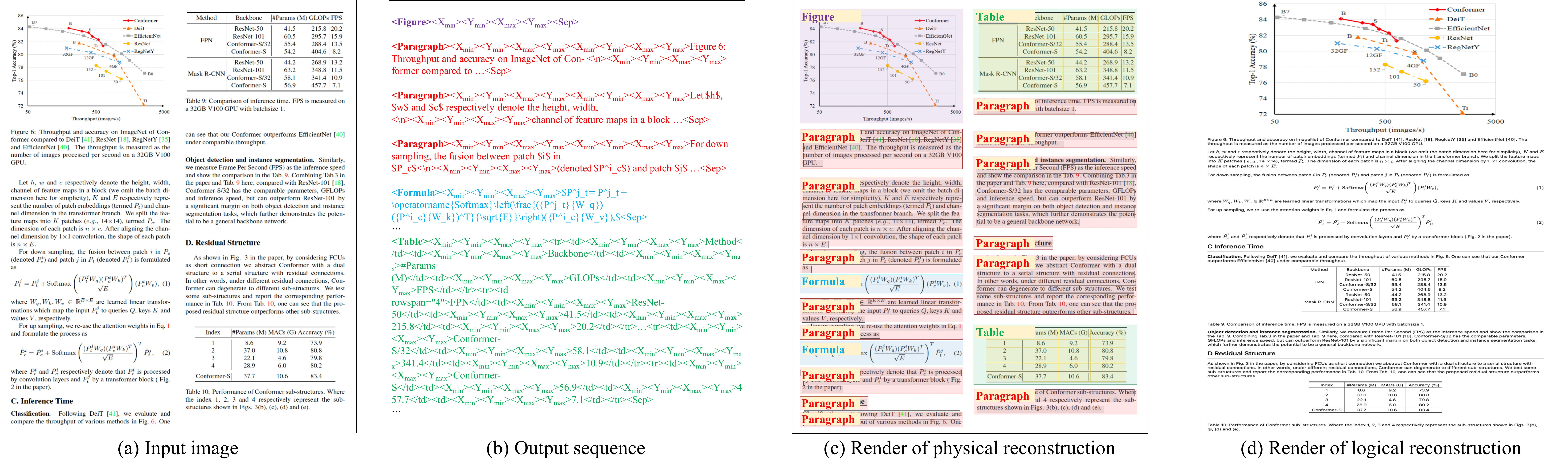}
	\caption{Visualization results of our proposed DREAM.}
	\label{fig:vis}
\end{figure*}
\noindent\textbf{Document reconstruction:}
In Tab.~\ref{tab:document_reconstruction}, we compare the performance of DREAM with the multi-stage based method~(\textit{i.e.}, PaddleOCR~\cite{du2020pp}) and the end-to-end methods~(\textit{i.e.}, Pix2Struct~\cite{lee2023pix2struct} and Nougat\textsubscript{}~\cite{blecher2023nougat}) on the newly proposed DocRec1K dataset.
Note that existing end-to-end methodologies are incapable of explicitly outputting the categories and coordinates of layout elements, thereby rendering a direct comparison with DREAM using the DSM metric unfeasible.
To circumvent this limitation, identical training datasets are employed to fairly optimize the Pix2Struct\textsubscript{base}$^{*}$~\cite{lee2023pix2struct} and Nougat\textsubscript{base}$^{*}$~\cite{blecher2023nougat} in accordance with the document reconstruction task for the comparison.
Consequently, DREAM surpasses Nougat\textsubscript{base}$^{*}$ by 6.2\%, thereby indicating a significant enhancement in the new task.
We also transform the ground truth of DocRec1K into the markdown format and utilize NED as the evaluation metric to juxtapose these models, which shows a similar improvement trend.
Sample results of DREAM are visualized in Fig.~\ref{fig:vis}, which contain the output sequence, render of physical reconstruction and render of logical reconstruction.
More visualizations are provided in the appendix.

\noindent\textbf{Document layout analysis:}
The comparison results on the task of document layout analysis are illustrated in Tab.~\ref{tab:document_layout_analysis}.
Specifically, DREAM decreases average mAP on DocBank~\cite{li2020docbank} and PubLayNet~\cite{zhong2019publaynet} benchmarks by round 1\% in comparison to the best VGT~\cite{da2023vision}, which shows that our model is competitive.

\noindent\textbf{Text recognition:}
The performance of DREAM is also evaluated in relation to other algorithms in the context of text recognition.
As illustrated in Tab.~\ref{tab:text_recognition}, DREAM achieves an average F1-score of 89.1\% on FUNSD~\cite{jaume2019funsd}, SROIE~\cite{huang2019icdar2019} and CORD~\cite{park2019cord} datasets surpassing KOSMOS-2.5~\cite{lv2023kosmos} by 2.1\%, which indicates its text recognition capacity.

\noindent\textbf{Table structure recognition:}
Tab.~\ref{tab:table_structure_recognition} presents the performance of our model on ICDAR-2013~\cite{gobel2013icdar}, SciTSR~\cite{chi2019complicated} and TableBank~\cite{li2020tablebank} datasets.
When juxtaposed with the robust baseline NCGM~\cite{liu2022neural}, which incorporates meticulously designed structural components for the task of table structure recognition, our DREAM exhibits a marginal decrease of 1-2\% across the three distinct datasets. 
Despite the complexity inherent in the task, DREAM manages to achieve commendable metrics without any specific adjustments.

\begin{table}
	\centering
	\setlength{\tabcolsep}{6.2mm}
	\begin{tabular}{l|c|c}
		\toprule[1.5pt]
		\multirow{2}{*}{\tabincell{c}{Method}}  
		&\multicolumn{2}{c}{{DocRec1K}}   \\
		\cline{2-3}
		& DSM &  NED \\
		\hline
		PaddleOCR~\cite{du2020pp} & 65.3 &  72.8  \\
		Pix2Struct\textsubscript{base}$^{*}$~\cite{lee2023pix2struct} & 81.7 &  86.4  \\
		Nougat\textsubscript{base}$^{*}$~\cite{blecher2023nougat} & 85.2 &  88.6  \\
		\hline
		DREAM& 91.4  & 92.7 \\
		\bottomrule[1.5pt]
	\end{tabular}
	\caption{Performance on document reconstruction task. }
	\label{tab:document_reconstruction}
\end{table}
\begin{table}
	\centering
	\setlength{\tabcolsep}{0.5mm}
	\begin{tabular}{l|c|c|c}
		\toprule[1.5pt]
		\multirow{2}{*}{\tabincell{c}{Method}}
		&\multicolumn{1}{c|}{{DocBank}}   
		&\multicolumn{1}{c|}{{PubLayNet}} 
		&\multicolumn{1}{c}{{DocLayNet}}  \\
		\cline{2-4}
		& mAP & mAP  & mAP\\
		\hline
		TransDLANet~\cite{cheng2023m6doc} & 68.4 & 94.5 & 72.3 \\
		LayoutLMv3\textsubscript{base}~\cite{huang2022layoutlmv3} & 78.3 & 95.1 & - \\
		VGT~\cite{da2023vision}  & 84.1 & 96.2 & - \\
		\hline
		DREAM& 83.2  & 95.3 & 74.8  \\
		\bottomrule[1.5pt]
	\end{tabular}
	\caption{Performance on document layout analysis task.}
	\label{tab:document_layout_analysis}
\end{table}
\begin{table}
	\setlength{\tabcolsep}{1.5mm}
	\centering
	\begin{tabular}{l|c|c|c}
		\toprule[1.5pt]
		\multirow{2}{*}{\tabincell{c}{Method}}  		
		&\multicolumn{1}{c|}{{FUNSD}} 
		&\multicolumn{1}{c|}{{SROIE}}  
		&\multicolumn{1}{c}{{CORD}}  \\
		\cline{2-4}
		&  F1-score & F1-score  & F1-score \\
		\hline
		Commercial OCR~\cite{lv2023kosmos} & 82.9 & 89.7& 84.3 \\
		KOSMOS-2.5~\cite{lv2023kosmos} & 83.3 & 92.1& 85.7 \\
		\hline
		DREAM& 86.5 & 93.7& 87.1 \\
		\bottomrule[1.5pt]
	\end{tabular}
	\caption{Performance on text recognition task.}
	\label{tab:text_recognition}
\end{table}
\begin{table}
	\setlength{\tabcolsep}{1.5mm}
	\centering
	\begin{tabular}{l|c|c|c}
		\toprule[1.5pt]
		\multirow{2}{*}{\tabincell{c}{Method}}  
		&\multicolumn{1}{c|}{{ICDAR-2013}} 
		&\multicolumn{1}{c|}{{SciTSR}} 
		&\multicolumn{1}{c}{{TableBank}}  \\
		\cline{2-4}
		& F1-score & F1-score & BLEU \\
		\hline
		FLAG-Net~\cite{liu2021show}& 98.6 & 99.5 & 93.9 \\
		TSRFormer~\cite{lin2022tsrformer}& - & 99.6 & - \\
		VAST~\cite{huang2023improving} & 96.5 & 99.5 & - \\
		NCGM~\cite{liu2022neural} & 99.6 & 99.7 & 94.6  \\
		\hline
		DREAM&  98.5 &  98.2 &  94.1 \\
		\bottomrule[1.5pt]
	\end{tabular}
	\caption{Performance on table structure recognition task.}
	\label{tab:table_structure_recognition}
\end{table}
\begin{table}
	\setlength{\tabcolsep}{1.5mm}
	\centering
	\begin{tabular}{l|c|c}
		\toprule[1.5pt]
		\multirow{2}{*}{\tabincell{c}{Method}}  
		&\multicolumn{1}{c|}{{IM2LATEX-100K}} 
		&\multicolumn{1}{c}{{CROHME-2014}} \\
		\cline{2-3}
		& BLEU & BLEU   \\
		\hline
		I2L-STRIPS~\cite{singh2018teaching}  & 88.9 & - \\
		IM2TEX~\cite{deng2017image} &  87.7 &  68.6 \\
		\hline
		DREAM& 89.4 & 66.3  \\
		\bottomrule[1.5pt]
	\end{tabular}
	\caption{Performance on formula recognition task.}
	\label{tab:formula_recognition}
\end{table}
\begin{table}
	\setlength{\tabcolsep}{8mm}
	\centering
	\begin{tabular}{l|c}
		\toprule[1.5pt]
		\multirow{2}{*}{\tabincell{c}{Method}}  
		&\multicolumn{1}{c}{{ReadingBank}}   \\
		\cline{2-2}
		& BLEU   \\
		\hline
		Heuristic Method~\cite{wang2021layoutreader} & 69.7 \\
		LayoutReader~\cite{wang2021layoutreader} & 98.2  \\
		TPP~\cite{zhang2023reading} & 98.2 \\
		DocReL~\cite{li2022relational} & 98.4 \\
		\hline
		DREAM&  {96.5}  \\
		\bottomrule[1.5pt]
	\end{tabular}
	\caption{Performance on reading order detection task.}
	\label{tab:reading_order_detection}
\end{table}

\noindent\textbf{Formula recognition:}
In the task of formula recognition, our model outperforms IM2TEX~\cite{deng2017image} on the IM2LATEX-100K~\cite{deng2017image} dataset, as evident in Tab.~\ref{tab:formula_recognition}.
However, the performance of DREAM on the handwritten dataset CROHME-2014~\cite{mouchere2014icfhr} is relatively lower than that of IM2TEX.
We attribute this lower accuracy primarily to the fact that our model is trained on synthetic data on rendered markup, rather than on datasets of handwritten expressions.

\noindent\textbf{Reading order detection:}
Tab.~\ref{tab:reading_order_detection} offers a comparative analysis of the impact of DREAM and other baselines on the ReadingBank~\cite{wang2021layoutreader} dataset in the task of reading order detection.
The BLEU score of DREAM is slightly lower than that of DocReL~\cite{li2022relational} by approximately 2\%. 
This discrepancy can be attributed to the fact that DocReL incorporates textual and positional information as input, whereas our model relies solely on image input.

\subsection{Ablation Study}
To better investigate the efficacy of various components within our proposed DREAM model, we conduct a comprehensive series of ablation studies in the document reconstruction task of which the results are summarized in Tab.~\ref{tab:ablation}.

\begin{table}[ht]
	\setlength{\tabcolsep}{1mm}
	\centering
	\begin{tabular}{l|cccc|cc}
		\toprule[1.5pt]
		
		Method & IE & FA& PD & NPD & DSM & NED \\
		\hline
		DREAM$^{*}$ w/o PD\&FA & \cmark& \xmark  & \xmark& \cmark & 84.8 &  88.1 \\
		DREAM$^{*}$ w/o PD & \cmark& \cmark  & \xmark& \cmark & 86.7 & 89.1 \\
		\hline
		\textbf{DREAM} & \cmark& \cmark  & \cmark& \xmark & \textbf{91.4} & \textbf{92.7} \\
		\bottomrule[1.5pt]
	\end{tabular}	
	\caption{Ablation studies of DREAM on DocRec1K dataset. ``IE'', ``FA'', ``PD'' and ``NPD'' stand for ``Image Encoder'', ``Feature Aggregator'', ``Parallel Decoder'' and ``Non-Parallel Decoder'' respectively. ``w/o'' is short for ``without''.}
	\label{tab:ablation}
\end{table}

\noindent\textbf{Effect of Feature Aggregator:}
Compared with the performance of the model designated as ``-w/o PD\&FA'', which is equipped with an image encoder and a text decoder akin to the Nougat, the introduction of FA~(denoted as ``-w/o PD'') to the model results in an enhancement of 1.9\% in DSM and 1.0\% in NED on DocRec1K dataset.
This confirms the advantage of query embeddings and further demonstrates that the supplementary tokens provided by FA can help DREAM more effectively capture the element representations within the document image.

\begin{figure}[t]
	\centering
	\includegraphics[width=1\linewidth]{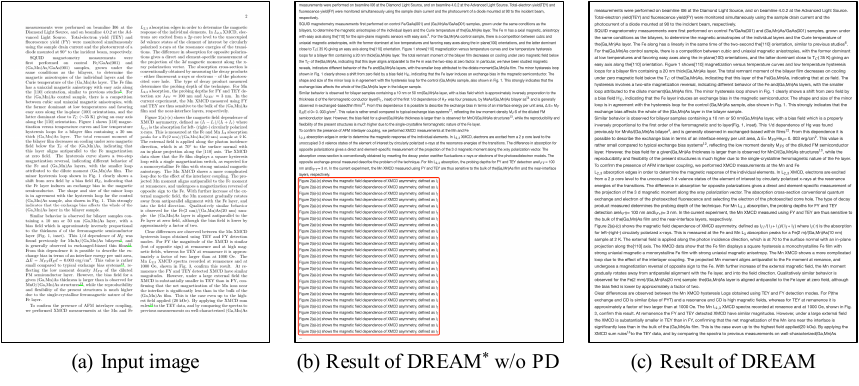}
	\caption{Visualizations between DREAM$^{*}$ w/o PD and DREAM.}
	\label{fig:vis2}
\end{figure}

\noindent\textbf{Effect of Parallel Decoder:}
As evidenced in Tab.~\ref{tab:ablation}, the performance on the DocRec1K dataset can be markedly augmented by incorporating PD into the DREAM model, as compared to the non-parallel version~(denoted as ``-w/o PD'').
This suggests that our parallel autoregression scheme is more proficient for the document reconstruction task.
To further investigate the working pattern of the parallel decoder, in Fig. ~\ref{fig:vis2}, we present a visualization of the predictions between the parallel and non-parallel versions.
It is intriguing to note that the model equipped with a non-parallel decoder tends to devolve into a repeating loop with an extended sequence, whereas the parallel version can significantly mitigate this phenomenon.
We hypothesize that the parallelization process curtails the length of the output sequence, thereby enhancing the stability of the model during the decoding process. 
Owing to the parallel decoder, our DREAM model can attain superior performance in the task of document reconstruction.

\subsection{Computational Complexity}
In order to conduct a comparative analysis of the computational complexity inherent in existing methodologies for document reconstruction, we summarize the model sizes and the inference time on the DocRec1K dataset.
in Tab.~\ref{tab:complexity}.
It is important to highlight that we have expanded the original Nougat~\cite{blecher2023nougat} model, which predicts markup texts, to the Nougat\textsubscript{base}$^{*}$version. 
This adaptation was made in accordance with the task proposed in Sec.~\ref{sec:definition}, thereby ensuring a fair comparison. 
When juxtaposed with the autoregressive model Nougat\textsubscript{base}$^{*}$, our DREAM model demonstrates a reduction in inference time cost with fewer parameters, which can be attributed to its parallel manner.
Despite DREAM showing promising potential in the realm of document reconstruction, the time-intensive nature of generative modeling remains a challenge. 
This issue necessitates further exploration in future research within related domains.

\begin{table}[htp]
	\setlength{\tabcolsep}{5.5mm}
	\centering
	\begin{tabular}{l|cc}
		\toprule[1.5pt]
		Method & \#Param& Inference Time \\
		\hline
		Nougat\textsubscript{base}$^{*}$~\cite{blecher2023nougat}  & 350 & 14.1  \\
		\hline
		DREAM & 270 & 3.2 \\ 
		\bottomrule[1.5pt]
	\end{tabular}
	\caption{Computational complexity comparison of different methods. ``\#Param'' denotes the number of parameters (M), while ``Inference Time'' means the time cost to run each image (s/image).}
	\label{tab:complexity}
\end{table}

%% file: 6_conclusion.tex
\section{Conclusion}
In this paper, we introduce a new task within the realm of document analysis and recognition, named document reconstruction, and present the DSM metric and DocRec1K dataset for assessing the efficacy of this task.
To facilitate the research aimed at this task, we propose a novel model, termed DREAM, to transform the document image into a sequence of document reconstruction in a holistic autoregressive manner, which encompasses more comprehensive information on document elements.
Extensive experiments on public benchmarks demonstrate its superiority over existing state-of-the-art methods.

%% file: X_suppl.tex
\begin{appendices}
\section{Multi-stage based Document Reconstruction}
The procedure of multi-stage document reconstruction~\cite{zhong2019publaynet, pfitzmann2022doclaynet, li2020docbank, cheng2023m6doc, liu2020abcnet, lyu2018mask, huang2022swintextspotter, jaume2019funsd, huang2019icdar2019, park2019cord, gobel2013icdar, chi2019complicated, zhong2020image, li2020tablebank, liu2021show, nassar2022tableformer, liu2022neural, huang2023improving, mouchere2014icfhr, deng2017image, yuan2022syntax, wang2021layoutreader} is traditionally divided into a pipeline of three stages, including 1) analyzing the layout elements within images, 2) arranging the elements in a sequential order, and 3) extracting the contents from diverse elements.

\noindent\textbf{Analyzing the layout elements within images:}
This process is referred to as layout analysis, which entails identifying and categorizing various components within a document image.
A dominant category of methodologies employed in document layout analysis incorporates classical vision tasks to discern the document elements.
For instance, conventional computer vision architectures such as the RCNN series~\cite{ren2015faster,  he2017mask}, YOLO series~\cite{jocher2021ultralytics} and DETR series~\cite{carion2020end, zhu2020deformable} are utilized for the detection or segmentation of layout objects.
Furthermore, a group of methods introduces unique designs for document layout analysis to enhance task performance.
TransDLANet~\cite{cheng2023m6doc} employs an adaptive element matching mechanism, enabling query embedding to better match ground truth to improve the effectiveness of layout analysis.
The method~\cite{li2020docbank} utilizes the sequence labeling models~\cite{devlin2018bert, liu2019roberta} to ascertain the semantic categories of disparate elements.
Moreover, a range of methods conceptualizes layout analysis as a multimodal task, integrating visual, layout and textual information to yield more accurate and comprehensive results.
The LayoutLM series~\cite{huang2022layoutlmv3} leverages self-supervised pre-training techniques within the transformer architecture to effectively represent documents.
A multi-modal model~\cite{barman2021combining} based on the U-Net architecture is proposed for segmenting layout elements, which amalgamates pixel-level visual features combined with text embedding maps generated by OCR output, subsequently producing pixel-level labels for layout elements.
Contrasting with the multimodal early fusion approach used in the previous model, MFCN~\cite{yang2017learning} implements a late fusion technique, incorporating text embedding maps at the end of the network.
Additionally, an auxiliary decoder is introduced for unsupervised image reconstruction to enhance the representation effectiveness of the encoder.
VSR~\cite{zhang2021vsr} utilizes a dual-stream convolutional network to extract visual and semantic features specific to each modality. Instead of merely concatenating or adding these features, a multi-scale adaptive aggregation module is designed to facilitate effective multimodal fusion.
Subsequently, a relation module is utilized to model the relationships among the generated set of layout component candidates.

\noindent\textbf{Arranging the elements in a sequential order:}
The task of sequencing necessitates the establishment of an order among the layout objects positioned on a page, with the objective of reflecting the manner in which humans process written information. 
A plethora of methodologies have been proposed in academic literature~\cite{cattoni1998geometric} to address this concern, employing either geometric or typographic attributes of the page objects. 
Some methodologies delve further, harnessing the content of the objects, irrespective of whether there is pre-existing knowledge of a specific document class or not.
The X-Y cut algorithm~\cite{jain1999data,ha1995recursive,ishitani2003document,meunier2005optimized} functions as a technique for page segmentation, capitalizing on the analysis of image pixel distribution.
This algorithm conducts a thorough examination of the distribution across both the horizontal~(X-axis) and vertical~(Y-axis) dimensions to identify continuous white spaces or gaps, which act as potential indicators of boundaries between text blocks or columns.
Blocks possessing a hierarchical structure are ultimately segmented from the page by recursively applying X-Y cuts.
In order to enhance performance on the task of reading order detection, LayoutReader~\cite{wang2021layoutreader} utilizes a seq2seq model to capture the word sequence, which can be intuitively understood by human readers.
Despite the proficiency of LayoutReader in restoring the reading order, it additionally integrates textual and positional information as input, as opposed to the original image.

\noindent\textbf{Extracting the contents from diverse elements:}
The process of extracting textual content from various document elements necessitates the application of a range of methodologies.
For plain text, text recognition techniques~\cite{liu2020abcnet, lyu2018mask, huang2022swintextspotter} are employed to convert texts within images into a format that is amenable to editing.
Mask TextSpotter~\cite{lyu2018mask} introduces an end-to-end trainable neural network specifically designed for scene text spotting.
ABCNet ~\cite{liu2020abcnet} innovatively adapts to fit oriented or curved text via a parameterized Bezier curve, thereby facilitating the recognition of text within images.
SwinTextSpotter~\cite{huang2022swintextspotter} incorporates a transformer encoder with the dynamic head as the detector, unifying the tasks of detection and recognition to leverage feature interaction.
In terms of the table content, methodologies~\cite{chi2019complicated, liu2021show, nassar2022tableformer, liu2022neural, huang2023improving} aimed at table structure recognition are utilized to identify the sub-elements such as rows and columns within cells, thereby enabling the extraction of both physical positions and logical contents.
The approach~\cite{chi2019complicated} employs DGCNN to predict the relationship between words represented by the appearance and geometry features.
FLAG-Net~\cite{liu2021show} harnesses the modulatable dense and sparse context of table elements in an end-to-end manner.
VAST~\cite{huang2023improving} introduces a sequential modeling framework to generate both logical and physical structures of tables.
Furthermore, formula recognition methodologies~\cite{deng2017image, yuan2022syntax} are utilized to handle mathematical content, which includes symbols, operators and variables.
IM2TEX~\cite{deng2017image} employs a neural encoder-decoder model to convert images into presentational markup based on a scalable coarse-to-fine attention mechanism.
SAN~\cite{yuan2022syntax} introduces a simple yet efficient method to incorporate syntax information into an encoder-decoder network.

\section{DocRec1K Dataset}
The DocRec1K dataset, as introduced in the main text, is a derivative of the DocLayNet~\cite{pfitzmann2022doclaynet} dataset and comprises 1,000 document images along with their respective annotations. 
This section offers an exhaustive elucidation of the ground-truth annotation processing, which is partitioned into three distinct phases: 1) the ordering of layout elements, 2) the correlation of these elements with their corresponding textual content, and 3) the transformation of element contents into the specific formats as previously outlined.

\noindent\textbf{Detecting the order of layout elements:}
The optimized xy-cut algorithm~\cite{meunier2005optimized} is employed to organize the pre-labeled layout elements, aligning them in accordance with the reading sequence.
During the sorting process, it is noted that the xy-cut technique occasionally fails to ascertain the sequence of certain layout elements due to overlapping bounding boxes, a consequence of labeling inaccuracies.
Consequently, the introduction of supplementary auxiliary sorting rules becomes necessary.
The frequently utilized top-to-bottom and left-to-right sorting rule presents a predicament: it is exceedingly sensitive to the y-coordinate of the layout elements, and even a minor alteration in the y-coordinate can lead to significantly divergent results. 
To alleviate this issue, a tolerance range is introduced during the top-to-bottom sorting process. 
If the y-coordinates of various layout elements reside within the same range, they are considered to occupy the same horizontal position and are subsequently arranged from left to right along the x-axis.

\noindent\textbf{Association the elements with corresponding content:}
DocLayNet~\cite{pfitzmann2022doclaynet} does not inherently create a direct association between the unformatted text content and the corresponding layout components.
To rectify this limitation, the Intersection-over-Union (IOU) criterion is strategically utilized to allocate each segment of unformatted text to a specific layout component. 
During this allocation process, there are instances where a single line of text is categorized under multiple text boxes.
Consequently, a consolidation process is implemented for multiple text boxes that fall under this classification.

\noindent\textbf{Transformation of element contents.}
The original annotated contents in DocLayNet~\cite{pfitzmann2022doclaynet} are not optimally suited for two specific types of components: tables and formulas.
To circumvent this constraint, the PaddleOCR~\cite{du2020pp} is employed to extract the tables and formulas within the document images.
More specifically, SLANet~\cite{li2022pp} is utilized to detect the coordinates of cell boxes and their corresponding row and column relationships, while CAN~\cite{li2022counting} is used to transcribe the formula image to the LaTeX sequence of mathematical expressions.

It is crucial to highlight that the pseudo-labeled contents, which include the order of elements, the structure of the table, and the information of the formula, are subjected to human intervention to validate the results and rectify any inaccuracies, thereby ensuring the quality of annotations.
Following these processes, the processed labels of various elements are compiled into a sequence of physical and logical information as described in Sec. 3.1 of the main text, which is denoted as the annotations for document reconstruction.

\section{Evaluation Datasets}
As introduced in the main text, the assessment is conducted on six distinct categories of tasks pertinent to document analysis and recognition: 1) The document reconstruction task on DocRec1K, as established in Sec.~3.3 of the main text, is evaluated using the DSM delineated in Sec.~3.2 of the main text and the widely accepted NED, 2) The document layout analysis task on DocBank~\cite{li2020docbank}, PubLayNet~\cite{zhong2019publaynet} and DocLayNet~\cite{pfitzmann2022doclaynet}, employs mAP as a measurement tool, 3) The text recognition task on FUNSD~\cite{jaume2019funsd}, SROIE~\cite{huang2019icdar2019} and CORD~\cite{park2019cord} utilizes the F1-score for evaluation, 4) The table structure recognition task on ICDAR-2013~\cite{gobel2013icdar}, SciTSR~\cite{chi2019complicated} and TableBank~\cite{li2020tablebank} leverages F1-score/BLEU as the evaluative metrics, 5) The formula recognition task on IM2LATEX-100K~\cite{deng2017image} and CROHME-2014~\cite{mouchere2014icfhr} is assessed under the BLEU protocol, and 6) The reading order detection task on ReadingBank~\cite{wang2021layoutreader} applies the BLEU.

\noindent\textbf{Datasets for document  reconstruction:}
The dataset, DocRec1K, comprising 1,000 document images, is introduced in this study to evaluate the efficacy of the document reconstruction task.
The performance of end-to-end and logical reconstruction is assessed using the DSM and NED respectively.

\noindent\textbf{Datasets for document layout analysis:} 
DocBank~\cite{li2020docbank}, a benchmark dataset specifically curated for document layout analysis, consists of 500K pages extracted from LaTeX documents accessible on arXiv.
This dataset is assembled employing a simple yet effective weak-supervised method, which utilizes unique colors to distinguish between different semantic structures, and subsequently generates annotations through a color-to-structure mapping process.
PubLayNet~\cite{zhong2019publaynet}, another dataset, includes over 360K document images that are publicly accessible on PubMed Central\textsuperscript{TM}. 
It is created by automatically aligning the XML representations with the content of PDF articles, thereby facilitating the annotation of common document layout elements. 
DocLayNet~\cite{pfitzmann2022doclaynet}, a dataset for document-layout annotation in COCO format, comprises over 80K manually annotated pages from a variety of data sources, representing a broad range of layout variations. 
For each PDF page in this dataset, the layout annotations provide labeled bounding boxes, with a selection of 11 unique classes.

\noindent\textbf{Datasets for text recognition:} 
We adopt the KOSMOS-2.5\cite{lv2023kosmos} approach to utilize the word-level F1-score to evaluate on FUNSD~\cite{jaume2019funsd}, SROIE~\cite{huang2019icdar2019} and CORD~\cite{park2019cord} datasets.
FUNSD~\cite{jaume2019funsd} comprises 199 real and scanned forms, which can be utilized for text detection, text recognition, spatial layout analysis, and entity linking.
SROIE~\cite{huang2019icdar2019} consists of 1,000 scanned receipt images, structured around three tasks: scanned receipt text localization, scanned receipt OCR, and key information extraction from scanned receipts.
CORD~\cite{park2019cord} contains 1,000 Indonesian receipts, in which the box/text annotations for OCR and multi-level semantic labels for parsing are included.

\noindent\textbf{Datasets for table structure recognition:} 
ICDAR-2013~\cite{gobel2013icdar}, an assemblage of hundreds of tables, is meticulously curated from PDFs discovered through Google search, with a particular emphasis on those possessing distinct bounding boxes and well-organized cells.
SciTSR~\cite{chi2019complicated}, a large-scale table structure recognition dataset comprising 15K tables extracted from scientific papers, has been introduced to facilitate research advancements in complex table scenarios, specifically targeting those containing cells spanning multiple rows and columns.
TableBank~\cite{li2020tablebank} comprises 417K high quality labeled tables, utilizing a 4-gram BLEU score as the evaluation metric with a singular reference.

\noindent\textbf{Datasets for formula recognition:} 
IM2LATEX-100K~\cite{deng2017image}, a substantial compilation of rendered real-world mathematical expressions gathered from published articles, is employed for the image-to-markup task predicated on the reconstruction of mathematical markup from rendered images.
CROHME-2014~\cite{mouchere2014icfhr} is a mathematical expression recognition dataset, which operates in a distinct domain from our rendered images and is specifically designed for stroke-based OCR.

\noindent\textbf{Datasets for reading order detection:} 
ReadingBank~\cite{wang2021layoutreader} serves as a benchmark dataset that incorporates reading order, text, and layout information for 500K document images, spanning a diverse array of document types.
The reading order, referred to as the word sequence, is extracted from the internal Office XML code of the DocX files.
Subsequently, the layout information is procured by aligning the bounding box with each word in the sequence, using a specially designed coloring scheme for PDF rendering.

\section{Post-processing}
In order to assess the efficacy of our model across a variety of tasks, we implement a streamlined post-processing approach on the resultant data produced by DREAM, thereby transforming them into the respective formats required for each task.

\noindent\textbf{Post-processing for document reconstruction:}
For the DSM evaluation metric, the original output of the model is directly utilized to gauge the performance of document reconstruction, without necessitating any specific processing.
To assess the quality of image-to-markdown generation using the widely recognized NED, the predicted sequence is transmuted into the markdown markup language format.
This involves the removal of physical reconstruction information, including the category $c$ and bounding box $b$.
Regarding the transcription content $t$, the content coordinates of the sub-elements are discarded.
Lastly, the $\textless {\rm Sep}\textgreater$ token at the end of the element is substituted by two ${\rm \backslash n}$ to concatenate a markdown sequence.

\noindent\textbf{Post-processing for document layout analysis:}
In order to evaluate the performance of document layout analysis using the mAP tool, the physical reconstruction results of each document element, including categories and coordinates, are preserved, while the transcription information is eliminated.

\noindent\textbf{Post-processing for text recognition:}
In terms of the text recognition task, the logical reconstruction of the element $\textless {\rm Paragraph} \textgreater$ and $\textless {\rm Table} \textgreater$ is employed to derive the pure text results, with the aim of calculating the word-level F1-score.
Note that the coordinates and HTML tags within these two types of elements are removed.

\noindent\textbf{Post-processing for table structure recognition:}
The transcribed content of the element $\textless {\rm Table} \textgreater$ is utilized to evaluate the task of table structure recognition, and subsequently transformed into the corresponding input format in accordance with the evaluation tools of F1-score and BLEU used in various datasets.

\noindent\textbf{Post-processing for formula recognition:}
Similarly, the transcription of the element $\textless {\rm Formula} \textgreater$ in LaTeX language is employed to calculate the score of formula recognition using the BLEU.

\noindent\textbf{Post-processing for reading order detection:}
Given that the ReadingBank~\cite{wang2021layoutreader} dataset utilizes plain text to evaluate the performance of reading order, we retain the text information of paragraphs and tables and concatenate them into a comprehensive text sequence.

\section{More Visualization Results}
\begin{figure*}[htp]
	\centering
	\includegraphics[width=1\linewidth]{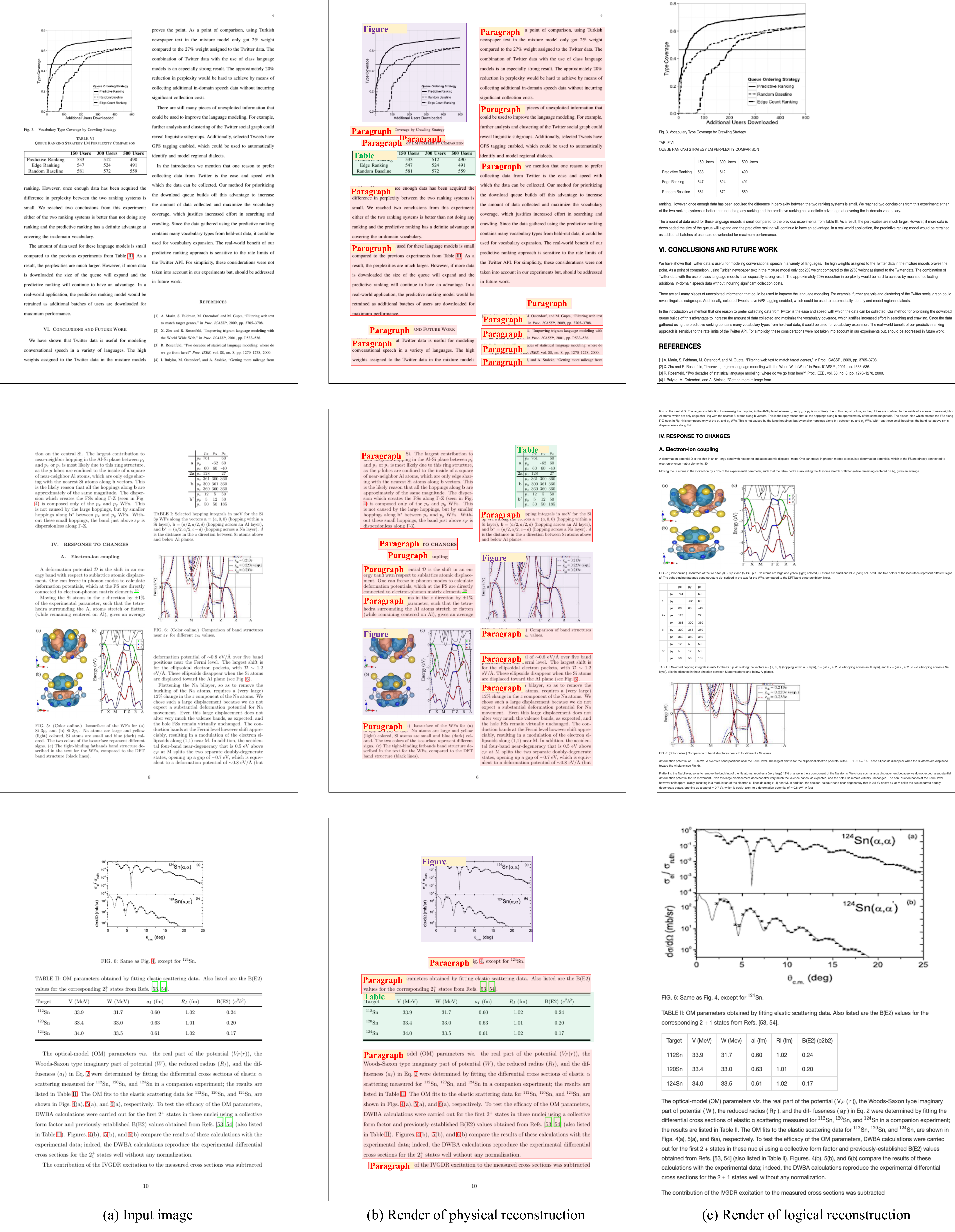}
	\caption{{More visualization results of DREAM.}} 
	\label{fig:app_goodcase}
\end{figure*}

Fig.~\ref{fig:app_goodcase} provides a comprehensive array of visualizations, encompassing the output physical reconstruction and logical reconstruction.
These depictions highlight the exceptional generalization proficiency of DREAM, demonstrating the capability to restore a broad spectrum of document contents in a more holistic manner, encompassing physical and logical reconstruction information.

However, it is apparent that our DREAM struggles with natural scene.
Future research endeavors will explore these challenges to further enhance the performance of document reconstruction.

\section{Broader Impact}
Recently, Large Language Models~(LLMs) have demonstrated significant potential owing to their text generation and comprehension abilities.
With an expanding spectrum of tasks and rising complexities, training with larger quantities of document image data, extending beyond mere texts, is essential for facilitating the performance of LLMs in the community.
Thanks to the potent effectiveness of DREAM aimed at the document reconstruction task, the document image can be effectively transformed into a sequence of diverse document elements, thereby facilitating the capacity of LLMs to interpret image data in the digital age.
This paper paves the way for a multitude of potential applications and opportunities in document analysis and understanding, advocating for researchers to improve the performance of document reconstruction in related fields.

\end{appendices}